\documentclass[final]{cvpr}

\usepackage{times}
\usepackage{epsfig}
\usepackage{graphicx}
\usepackage{amsmath}
\usepackage{amssymb}

\usepackage[utf8]{inputenc} %
\usepackage[T1]{fontenc}    %
\usepackage{url}            %
\usepackage{booktabs}       %
\usepackage{amsfonts}       %
\usepackage{nicefrac}       %
\usepackage{microtype}      %
\usepackage{capt-of}
\usepackage{cuted}

\usepackage[usenames, dvipsnames]{color}

\usepackage{multirow}
\usepackage{pifont}%
\newcommand{\xmark}{\ding{55}}%
\usepackage[font=small,labelfont=footnotesize]{caption}

\usepackage[pagebackref=true,breaklinks=true,colorlinks,bookmarks=false]{hyperref}

\def\cvprPaperID{5491} %
\def\confYear{CVPR 2021}

\begin{document}

\title{Learning to Segment Rigid Motions from Two Frames}

\author{Gengshan Yang$^{1\thanks{Code is available at \href{https://github.com/gengshan-y/rigidmask}{github.com/gengshan-y/rigidmask}.}}$,\quad Deva Ramanan$^{1,2}$\\
$^1$Carnegie Mellon University,\quad $^2$Argo AI\\
{\tt\small \{gengshay,\,deva\}@cs.cmu.edu}}

\maketitle

\begin{strip}\centering
\vspace{-40pt}
\includegraphics[trim={0cm 0cm 0cm 0cm},clip,width=\textwidth]{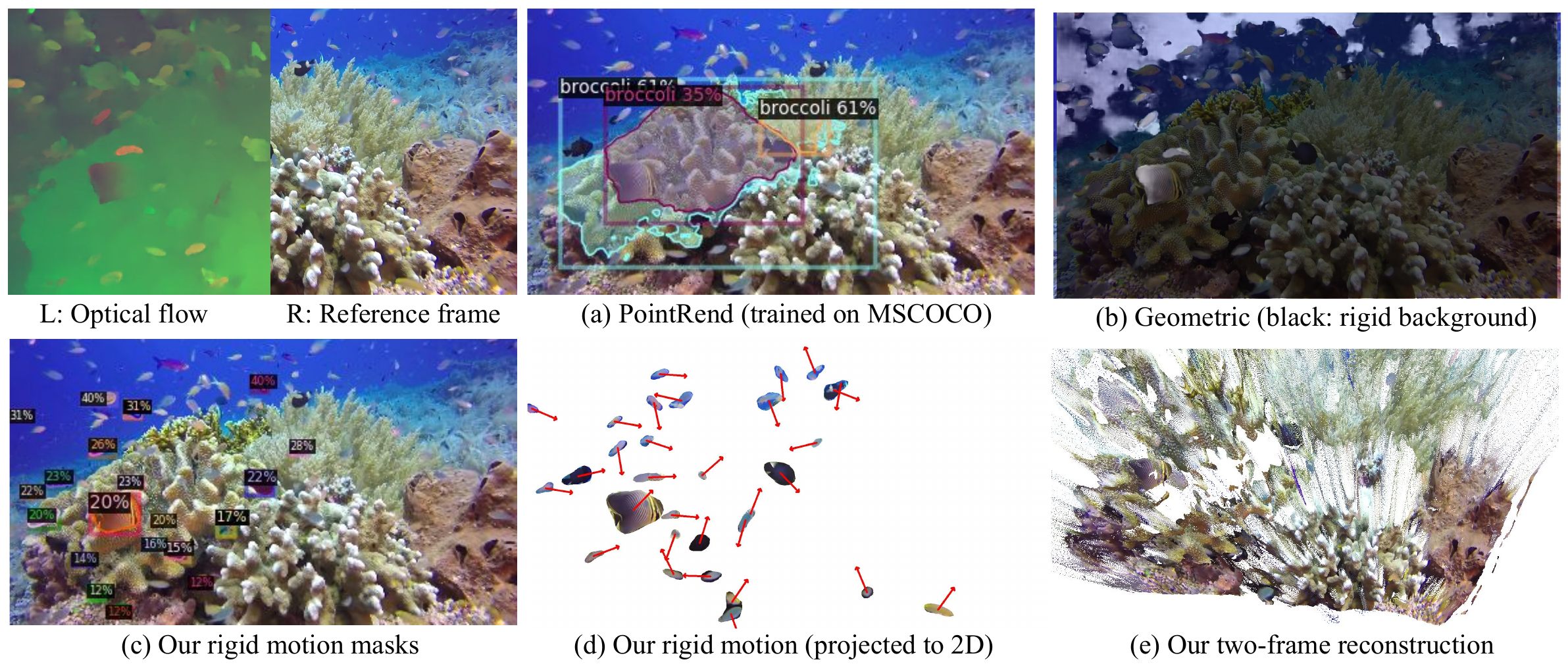}
\captionof{figure}{(a) Many data-driven segmentation methods heavily rely on appearance cues, and fail for novel test scenes. For instance, PointRend~\cite{kirillov2019pointrend} trained on MSCOCO fails to detect coral reef fishes even with a low confidence threshold of 0.1. (b) On the other hand, geometric motion segmentation~\cite{bideau2016s, Wulff:CVPR:2017} generalizes to novel appearance, but fails due to noisy flow inputs and degenerate motion configurations. (c)-(e) We propose a neural architecture powered by geometric reasoning that decomposes a scene into a rigid background and multiple moving rigid bodies, parameterized by 3D rigid transformations. It demonstrates generalization ability to novel scenes and robustness to noisy inputs as well as motion degeneracies. The inferred rigid motions significantly improve depth and scene flow accuracy.}
\label{fig:scooter}
\end{strip}

\begin{abstract}
 Appearance-based detectors achieve remarkable performance on common scenes, benefiting from high-capacity models and massive annotated data, but tend to fail for scenarios lack of training data. Geometric motion segmentation algorithms, however, generalize to novel scenes, but have yet to achieve comparable performance to appearance-based ones, due to noisy motion estimations and degenerate motion configurations. To combine the best of both worlds, we propose a modular network, whose architecture is motivated by a geometric analysis of what independent object motions can be recovered from an egomotion field. It takes two consecutive frames as input and predicts segmentation masks for the background and multiple rigidly moving objects, which are then parameterized by 3D rigid transformations. Our method achieves state-of-the-art performance for rigid motion segmentation on KITTI and Sintel. The inferred rigid motions lead to a significant improvement for depth and scene flow estimation. At the time of submission, our method ranked 1st on KITTI scene flow leaderboard, out-performing the best published method (SF error: 4.89\% vs 6.31\%).
\end{abstract}

\section{Introduction}
Autonomous agents such as self-driving cars need to be able to navigate safely in dynamic environments. Static environments are far easier to process because one can make use of geometric constraints (SFM/SLAM) to infer scene structure~\cite{durrant2006simultaneous}. Dynamic environments require the fundamental ability to both segment moving obstacles and estimate their depth and speed~\cite{barsan2018robust}. Popular solutions include object detection or semantic segmentation~\cite{lin2018architectural}. While one can build accurate detectors for many categories of objects that are able to move, ``being able to move'' is not equivalent to ``moving''. For example, there is a profound difference between a parked car and an ever-so-slightly moving car (that is pulling out of parked location), in terms of the appropriate response needed from a nearby autonomous agent. Secondly, class-specific detectors rely heavily on appearance cues and categories present in a training set. Consider a trash can that falls on the street; current {\em closed-world} detectors will likely not be able to model all types of moving debris. This poses severe implications for safety in the open-world that a truly autonomous agent must operate~\cite{bendale2015towards}.

\begin{figure}
\centering
\includegraphics[width=\linewidth]{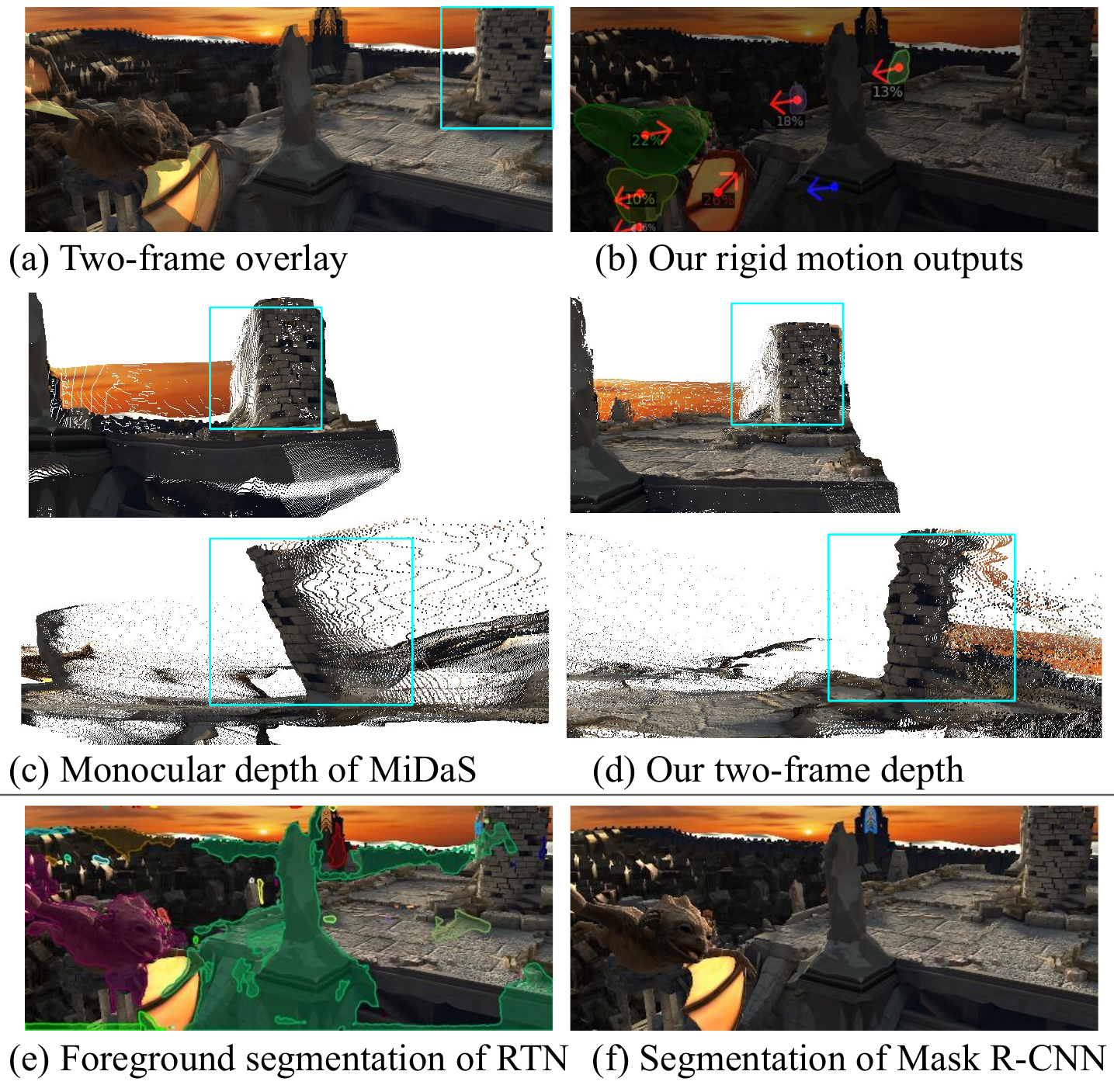}
\vspace{-15pt}
\caption{Results on Sintel sequence temple\_2, frame 17-18. (a)-(b) Our method segments rigid motions and fits 3D rigid transformations over two frames. The blue and red arrows indicate the estimated motion of the rigid background and parts respectively. (c)-(d) An initial depth is refined by triangulating optical flow within each rigid motion mask. Note that the tower in the cyan rectangle is leaning in the initial MiDaS~\cite{Ranftl2019} depth, but ``rectified'' by our method. (e)-(f) Our method segments rigid objects more reliably than the prior two-frame rigidity estimation method~\cite{Lv18eccv} and generalizes to novel appearance compared to appearance-based detectors~\cite{he2017mask}.}
\label{fig:moseg-sintel}
\vspace{-10pt}
\end{figure}

\noindent{\bf Problem formulation:} We follow historic work on motion-based perceptual grouping~\cite{irani1998unified,shi1998motion,tron2007benchmark,weber1997rigid,xu20193d} and segment moving objects without relying on appearance cues.
Specifically, we focus on segmenting {\em rigid} bodies from {\em two frames}. We focus on two-frame because it is the minimal set of inputs to study the problem of motion segmentation, and in practice, perception-for-autonomy needs to respond immediately to dynamic scenes, e.g., an animal that appears from behind an occlusion. We focus on rigid body and its {\em 3D} motion parameterizations because it's directly relevant for an autonomous agent acting in a 3D world. While dynamic scenes often contain nonrigid objects such as people, we expect that deformable objects may be modeled as a rigid body over short time scales, or decomposed into rigidly-moving parts~\cite{agin1976computer,biederman1993geon}. One example of our method decomposing a flying dragon into multiple rigid parts is shown in Fig.~\ref{fig:moseg-sintel}.

\noindent{\bf Challenges:} Earlier work on rigid motion segmentation often makes use of geometric constraints arising from epipolar geometry and rigid transformations. However, there are several fundamental difficulties that plague geometric motion segmentation. First, epipolar constraints fail when camera motion is close to zero~\cite{xu20193d}. Second, points moving along epipolar lines cannot be distinguished from the rigid background~\cite{yuan2007detecting}, which we discuss at length in Sec.~\ref{sec:geo-moseg}. Third, geometric criteria are often not robust enough to noisy motion correspondences and camera egomotion estimates, which can lead to catastrophic failures in practice.

\begin{figure*}
    \centering
    \includegraphics[width=\linewidth]{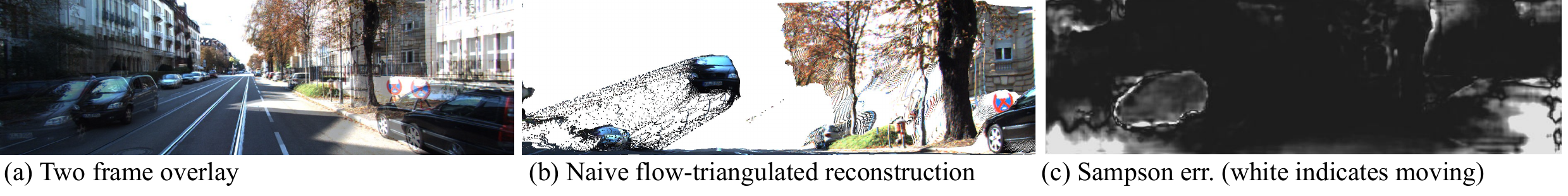}
    \vspace{-15pt}
	\caption{Collinear motion ambiguity. (a) The input scene contains a dynamic object (the car in the lower left) moving parallel to camera translational direction. (b) One can triangulate motion correspondences assuming overall \textit{rigidity} that places the moving car at an elevated height, which illustrates both (1) the commonality of this degenerate case~\cite{yuan2007detecting} in urban navigation, and (2) one solution of using structural scene priors that do not allow for floating objects above the ground. (c) Due to such ambiguities, the 2D motion of the moving car is \textit{consistent} with the camera egomotion, leaving it indistinguishable under classic motion segmentation metrics such as Sampson error~\cite{hartley2003multiple}.}
	\vspace{-5pt}
    \label{fig:ambiguity-collinear}
    \end{figure*}
    
\noindent{\bf Method:} We theoretically analyze ambiguities in 3D rigid motion segmentation, and resolve such ambiguities by exploiting recent techniques for upgrading 2D motion observation to 3D with optical expansion~\cite{yang2020upgrading} and monocular depth cues~\cite{Ranftl2019}. To deal with noisy motion correspondences and degenerate scene motion, we design a convolutional architecture that segments the rigid background and an arbitrary number of rigid bodies from a given motion field. Finally, we parameterize the 3D motion of individual rigid bodies by fitting 3D rigid transformations.

\noindent{\bf Contributions:} (1) We provide a geometric analysis for ambiguities in 3D rigid motion segmentation from 2D motion fields, and introduce solutions to deal with such ambiguities. (2) We propose a geometry-aware architecture for 3D rigid motion segmentation from two RGB frames, which is generalizable to novel appearance, resilient to different motion types and robust to noisy motion observations. (3) Our method achieves state-of-the-art (SOTA) performance of rigid motion segmentation on KITTI/Sintel. The inferred rigidity masks significantly improve the performance of downstream depth and scene flow estimation tasks.

\section{Related Work}
\noindent{\bf Geometric Motion Segmentation:}\quad The problem of clustering motion correspondences into groups that follow a similar 3D motion model has been extensively studied in the past~\cite{torr1998geometric,torr1999problem,tron2007benchmark,vidal2006two,vidal2003optimal,xu20193d,yuan2007detecting}. However, prior methods either focus on theoretical analysis with noisy-free data, or assume relatively simple scenes where long-term motion trajectories can be obtained by point tracking algorithms. Some recent work~\cite{bideau2016s, bideauCVPR18, Wulff:CVPR:2017} tackles more complex scenarios with two-frame optical flow inputs, where geometric constraints, such as motion angle and plane plus parallax (P+P)~\cite{sawhney19943d} are considered as cues of ``moving versus static''.
However, such geometric constraints are sensitive to noise in optical flow even under a robust estimation framework~\cite{fischler1981random}. Moreover, as we shall see in Sec.~\ref{sec:geo-moseg}, the prior two-frame solutions do not deal with several degenerate cases, including co-planar/co-linear motion~\cite{yuan2007detecting} and camera motion degeneracy~\cite{torr1999problem}. We address these problems by encoding geometric constraints into a modular neural network.

\noindent{\bf Learning-Based Video Object Segmentation:}\quad
Segmenting salient objects from videos historically stems from the problem of image salient object detection~\cite{ochs2013segmentation, Perazzi2016}, where existing methods often rely either on appearance features or on salient motions from 2D optical flow~\cite{jain2017fusionseg,lu2019see,tokmakov2017learning,tokmakov2019learning, yang2019anchor,zhou2020motion}. Oftentimes, optical flow is interpreted as a color image~\cite{jain2017fusionseg,zhou2020motion}, where geometric information, such as camera egomotion, is ignored. Close to our methodology, Motion Angle Network (MoA-Net)~\cite{bideau2018moa}, analytically reduces the effect of camera rotation and uses the ``rectified'' flow angle as input features to a binary segmentation network. Our approach further incorporates 3D flow and depth cues and segments multiple 3D rigid motions.

\noindent{\bf Instance Scene Flow:}\quad
Scene flow is the problem of resolving dense 3D scene motion from an ego-camera%
~\cite{Menze2015CVPR,vedula1999three}, which is inherently challenging due to the lack of visual evidence for correspondence matching in the local patch, for example, occluded or specular surfaces. Prior approaches often utilize scene rigidity priors to resolve such ambiguities, such as piecewise rigidity prior~\cite{Menze2015CVPR, Vogel2015IJCV} and semantic rigidity prior~\cite{Behl2017ICCV,Ma2019CVPR}. Then, scene flow is parameterized as the 3D motion of each rigid segment. However, it is risky to segment the scene purely relying on semantics -- object that is able to move is not the same as objects that are moving. Furthermore, such high-level cues do not generalize to an open-world, where algorithms are required to be robust to never-before-seen categories~\cite{bendale2015towards}. Instead, we exploit \emph{motion rigidity} for scene flow estimation, which decomposes the scene into multiple rigidly moving segments while preserving the completeness of individual rigid bodies.

\begin{figure*}
\centering
\includegraphics[width=\linewidth]{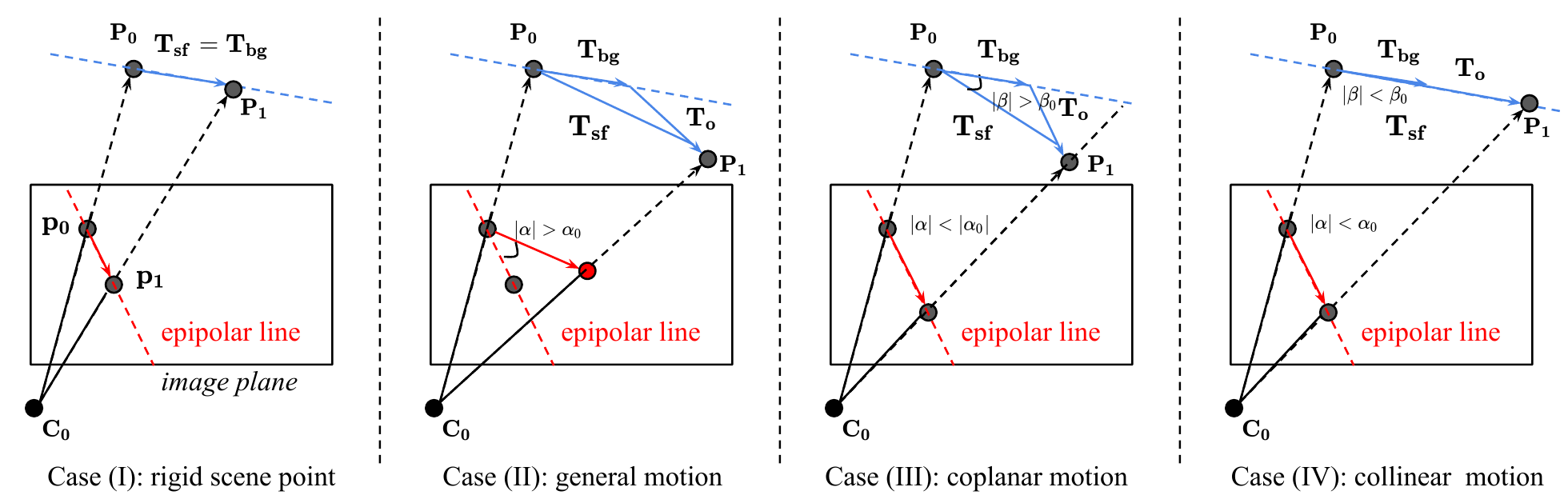}
 \vspace{-15pt}
\caption{When can a scene point ${\bf P}$ be identified as moving? Assuming camera rotation has been removed, we have 3D scene flow (defined as motion relative to the camera) ${\bf T_{sf}}={\bf T_{bg}}+{\bf T_{o}}$, where ${\bf T_{bg}}=-{\bf T_{c}}$ is the rigid background motion induced by the camera motion, and ${\bf T_{o}}$ is the independent object motion. Case (I): Assuming a rigid scene point ${\bf P}$ with zero independent motion ${\bf T_o}= {\bf 0}$, the 2D motion projected by ${\bf T_{sf}}$ must lie on the epipolar line. Case (II): In other words, if the 2D motion deviates from the epipolar line, $|{\alpha}|>{\alpha}_0$, ${\bf P}$ must be moving, analagous to Sampson error~\cite{hartley2003multiple}. Case (III): {\em However, the inverse does not hold.} If 2D flow is consistent with the background motion ($|{\alpha}|<{\alpha}_0$), ${\bf P}$ might still be moving in the epipolar plane. However, if the {\em angular direction} of 3D flow ${\bf T_{sf}}$ -- computable from optical expansion~\cite{yang2020upgrading} -- differs from ${\bf T_{bg}}$ ($|{\beta}|>{\beta}_0$), ${\bf P}$ must be moving. Case (IV): If the 3D flow direction is consistent with background motion ($|{\beta}|<{\beta}_0$), ${\bf P}$ could still be moving in the direction of ${\bf T_{bg}}$, making it unrecoverable without knowing the scale (or relative depth).
}
\label{fig:ambiguity}
\vspace{-5pt}
\end{figure*}

\section{Approach}

In this section, we first analyze degeneracies in motion segmentation that arise when dynamic motion is indistinguishable from the camera motion, and what information is required to resolve the ambiguities. We then design a neural architecture for rigid instance motion segmentation that builds on this geometric analysis, producing a pipeline for two-frame rigid motion segmentation.

\subsection{Two-Frame 3D Motion Segmentation}
\label{sec:geo-moseg}
\noindent{\bf Problem setup:}\quad Given two-frame motion correspondences $({\bf p_0},{\bf p_1}) \in \mathbb{R}^4$, observed by a monocular camera with known camera intrinsics $({\bf K_0},{\bf K_1})$, we are interested in detecting points whose 3D motion cannot be described by the camera motion ${\bf R_c} \in {\bf SO}(3)$, ${\bf T_c} \in \mathbb{R}^3$, such that
\begin{equation}\footnotesize
\label{eq:rigid-transform}
({\bf R_c}{\bf P_1} +{\bf T_c})-{\bf P_0} \neq {\bf 0}, %
\qquad \text{(transform of coordinates)}
\end{equation}
where ${{\bf P_0}=Z_0{\bf K_0}^{-1}{\bf \tilde{p}_0}}$ and ${\bf P_1}=Z_1{\bf K_1}^{-1}{\bf \tilde{p}_1}$ are corresponding 3D points observed in the camera coordinate systems of both frames, given corresponding depth $(Z_0,Z_1)$ and homogeneous coordinates $({\bf \tilde{p}_0},{\bf \tilde{p}_1})$. To gain more geometric insights, we re-arrange Eq.~\eqref{eq:rigid-transform} into
\begin{equation}\footnotesize
\begin{split}
\label{eq:2D-rt}
&{\bf T_{sf}} = {\bf K_0}^{-1}(Z_1{\bf H_R}{\bf \tilde{p}_1} -Z_0{\bf \tilde{p}_0}) \neq -{\bf T_c} , \\& (\text{``rectified'' 3D scene flow $\neq$ negative camera translation})
\end{split}
\end{equation}
where ${\bf T_{sf}}={\bf R_c}{\bf P_1} -{\bf P_0}$ is the ``rectified'' 3D scene flow, with the motion induced by camera rotation ${\bf R_c}$ removed through ``rectifying'' the second frame coordinate system to have the same orientation as the first frame; and ${\bf H_R}={\bf K_0}{\bf R_c}{\bf K_1}^{-1}$ is the rotational homography that “rectifies” the second image plane into the same orientation as the first image plane, removing the effect of camera
rotation from the 2D motion fields. Eq.~(\ref{eq:2D-rt}) states that the rectified 3D scene flow of a moving point ${\bf P}$ will not equal the negative camera translation. However, assuming the camera motion is known, there are still two crucial degrees of freedom that are undetermined: depth $Z_0$ and $Z_1$.

\noindent{\bf Coplanar motion degeneracy:}\quad
Solving for $Z_0$ and $Z_1$ equates to estimating the depth and 3D scene flow, which itself is challenging~\cite{Menze2015CVPR}. To remove such dependencies, classic geometric motion segmentation segments points whose \emph{2D motion} is inconsistent with the camera motion, measured either by Sampson distance to the epipolar line~\cite{hartley2003multiple, torr1998geometric} or plane plus parallax (P+P)~\cite{sawhney19943d} representations that factor out camera rotation, allowing one to evaluate the angular deviation of the 2D motion to the epipole~\cite{bideau2016s, irani1998unified}. 
However, {\em is 2D motion sufficient to segment points moving in 3D?} The answer is no (Fig.~\ref{fig:ambiguity-collinear}). Formally, 3D points that translate within the epipolar plane defined by the camera translation vector ${\bf T}_c$ will project to the epipolar line, making them "appear" as stationary points, as shown in Fig.~\ref{fig:ambiguity} Case (II).

To detect such co-planar motion, we make use of optical expansion cues that upgrade 2D flow to 3D as suggested by recent work~\cite{yang2020upgrading}. Optical expansion, measured by the scale change of overlapping image patches, approximates the relative depth $\tau=\frac{Z_1}{Z_0}$ for non-rotating scene elements under scaled orthographic projection~\cite{yang2020upgrading}. We derive a 3D motion angle criterion that does not require depth, but removes the ambiguity of points moving within the epipolar plane. Normalizing Eq.~(\ref{eq:2D-rt}) by depth $Z_0$, we have
\begin{equation}\footnotesize
\begin{split}
\label{eq:nsf-rt}
&{\bf \tilde{T}_{sf}} = {\bf K_0}^{-1}(\tau{\bf H_R}{\bf \tilde{p}_1} -{\bf \tilde{p}_0}) \not\sim -{\bf T_{c}}, 
\\ &(\text{rectified 3D flow direction $\neq$ neg. camera translation direction})
\end{split}
\end{equation}
where ${\bf \tilde{T}_{sf}}=\frac{{\bf {T}_{sf}}}{Z_0}$ is the rectified and normalized 3D flow and $ \not\sim$ indicates two vectors are different in their directions. Eq.~(\ref{eq:nsf-rt}) states that a point is moving if the direction of its rectified 3D scene flow is not consistent with the direction of the camera translation, as shown in Fig.~\ref{fig:ambiguity} Case (III).

\noindent{\bf Collinear motion degeneracy:}\quad
However, there is still a remaining ambiguity that cannot be resolved. If point ${\bf P}$ moves in the opposite direction of the camera translation, both classic criteria and Eq.~(\ref{eq:nsf-rt}) would fail, as shown in Fig.~\ref{fig:ambiguity} Case (IV). Such ambiguity remains even given multiple frames~\cite{yuan2007detecting}, but is common in many real-world applications, e.g., two cars passing each other (Fig.~\ref{fig:ambiguity-collinear}). To identify moving points in such cases, one could use depth $Z_0$ to recover the metric scale of normalized rectified scene flow ${\bf \tilde{T}_{sf}}$, and compare it with camera translation ${\bf T_c}$. However, in a monocular setup, we neither know the scale of ${\bf T_c}$ nor trust the overall scale of $Z_0$~\cite{hartley2003multiple}. Instead, we derive a depth contrast criterion, inspired by an observation that \textit{dynamic} scene points triangulated from flow assuming overall rigidity will appear ``abnormal'' in the 3D reconstruction, such as the floating car in Fig.~\ref{fig:ambiguity-collinear} (b). To do so, we contrast the flow-derived depth $Z^{flow}_{0}$ with a depth prior $Z_0^{prior}$,
\begin{equation}\footnotesize
\label{eq:depth-rt}
Z_0^{flow} \neq \gamma Z^{prior}_{0}, \quad (\text{flow-triangulated depth $\neq$ depth prior})
\end{equation}
where $Z^{flow}_{0}$ can be computed efficiently using midpoint or DLT triangulation algorithm~\cite{hartley2003multiple}, depth prior $Z^{prior}_{0}$ can be represented by a data-driven monocular depth network, and the scale factor $\gamma$ that globally aligns $Z^{prior}_{0}$ to $Z^{flow}_{0}$ can be determined by robust least squares~\cite{sun2010secrets}.

\noindent{\bf Egomotion degeneracy:}\quad
Furthermore, when the camera translation is small, ${\bf T_c}$ is notoriously difficult to estimate due to small motion parallax. In such cases, rigid-background motion (and objects that deviate from it) is easier to model with a rotational homography model~\cite{torr1998geometric,torr1998robust}. 

\begin{figure*}
\centering
\includegraphics[width=0.9\linewidth]{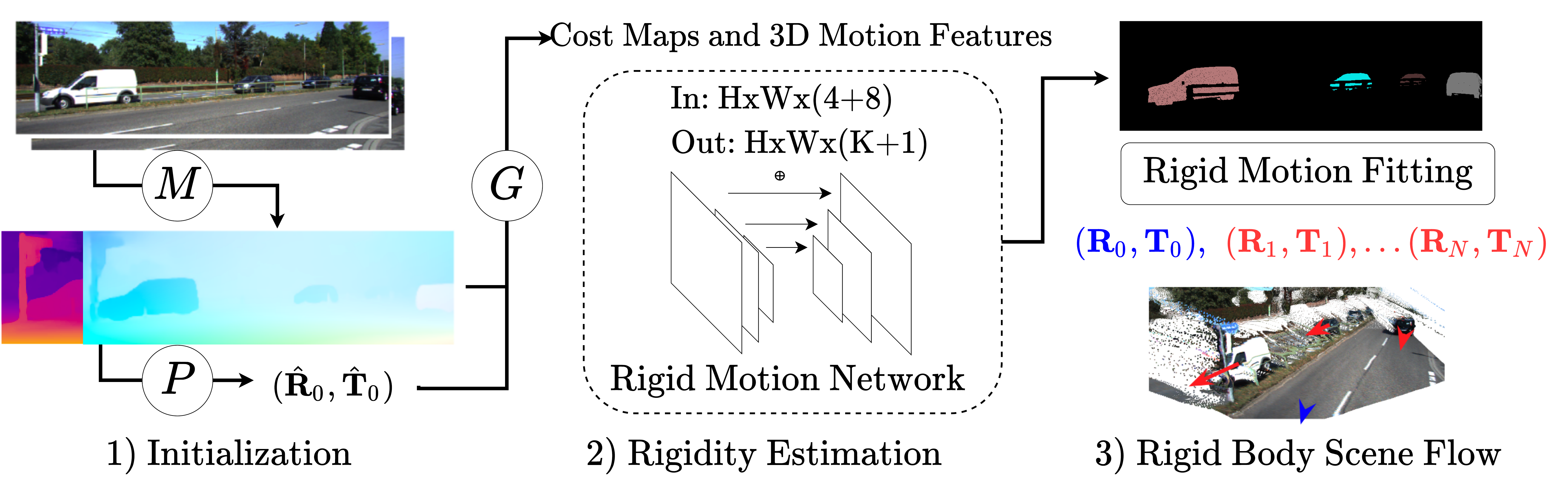}
\vspace{-10pt}
\caption{We detect and estimate rigid motions in three steps: First, initial 3D scene flow is computed using off-the-shelf networks (M) and camera motion is estimated by epipolar geometry (P) given two frames. Then, rigidity cost maps and rectified scene flow are computed (G) and fed into a two-stream network that produces the segmentation masks of a rigid background and an arbitrary number of rigidly moving instances. Finally, we fit a rigid transformation for the \textcolor{blue}{background} and each \textcolor{red}{rigid instance} to update their depth and 3D scene flow.}
\label{fig:arch}
\vspace{-10pt}
\end{figure*}

\subsection{Learning to Segment Rigid Motions}
We now operationalize our motion analysis into a deep network for rigid motion segmentation (Fig.~\ref{fig:arch}). At its heart, {\em our network learns to transform motion measurements (noisy 3D scene flow) into pixel-level masks of rigid background and instances.} To do so, we construct motion cost maps designed to address the motion degeneracies described earlier. Given such input maps and raw scene flow measurements, we use a two-stream network architecture that separately regresses the rigid background and rigid instance masks.

\noindent{\bf Motion estimation:}\quad
First, we extract the camera and relative scene motion given two frames. We apply existing methods to estimate optical flow, optical expansion and monocular depth~\cite{Ranftl2019,yang2020upgrading}. To estimate camera motion, we fit and decompose essential matrices from flow maps using the five-point algorithm with a differentiable and parallel RANSAC~\cite{brachmann2019ngransac}. 

\noindent{\bf Rigidity cost-maps inputs:}\quad
We construct rigidity cost maps tailored to camera-object motion configurations analysed in Sec.~\ref{sec:geo-moseg}, including (1) an epipolar cost for general configurations, computed as per-pixel Sampson error~\cite{hartley2003multiple}; (2) a homography cost to deal with small camera translations, implemented as per-pixel reprojection error according to the rotational homography~\cite{dubrofsky2009homography} in Eq.~(\ref{eq:2D-rt}); (3) a 3D P+P cost for coplanar motion configurations computed as Eq.~(\ref{eq:nsf-rt}); and (4) a depth contrast cost to deal with colinear motion ambituity, computed as Eq.~(\ref{eq:depth-rt}). 

\noindent{\bf Rectified scene flow inputs:}\quad
Besides rigidity cost-maps, we find it helpful to also input raw scene flow measurements, represented as an 8-channel feature map, containing the first frame 3D scene points ${\bf P_0}$, rectified motion fields ${\bf T_{sf}}$, and uncertainty estimations of flow and optical expansion $(\sigma_1,\sigma_2)$. To compute ${\bf P_0}$, we back-project the first frame pixel coordinates given monocular depth inputs; to compute ${\bf T_{sf}}$, we upgrade optical flow using optical expansion $\tau$, 
\begin{equation}\footnotesize
\begin{split}
\label{eq:rsf}
&{\bf \tilde{T}_{sf}} = {\bf K_0}^{-1}(\tau{\bf H_R}{\bf \tilde{p}_1} -{\bf \tilde{p}_0}),
\end{split}
\end{equation}
where the second coordinate frame is rectified by rotational homography ${\bf H_R}={\bf K_0}{\bf R_c}{\bf K_1}^{-1}$ to remove the effect of camera rotation. Finally, the uncertainty of optical flow and optical expansion are computed as out-of-range confidence score and Gaussian variance respectively~\cite{ilg2018uncertainty, yang2019volumetric}. Such rectified scene flow inputs are more effective than 2D optical flow, as empirically tested in ablation study (Tab.~\ref{tab:aba}).

\noindent{\bf Architecture and losses:}\quad
We use a two-stream architecture: (1) a lightweight U-Net~\cite{ronneberger2015u} architecture to predict binary labels for pixels belonging to the (rigid) background and (2) a CenterNet~\cite{zhou2019objects} architecture to predict pixel instance masks. Inspired by the single-shot segmentation head proposed in PolarMask~\cite{xie2020polarmask}, stream (2) outputs a heatmap representing object centers and a regression map of $K=36$ polar distances at evenly distributed angles. The overall architecture is end-to-end differentiable and can be trained with standard loss functions,
\begin{align}
L = \alpha_1 L_{\text{binary}} + \alpha_2 L_{\text{center}} + \alpha_3 L_{\text{polar}}
\end{align}
where $L_{\text{binary}}$ is binary cross-entropy loss with label balancing, $L_{\text{center}}$ is the focal loss and $L_{\text{polar}}$ is an $L_1$ regression loss evaluated at each ground-truth center location. Weights are empirically chosen as $\alpha_1 = 1^{-3}$, $\alpha_2 = 1^{-3}$ and $\alpha_3 = 1^{-8}$.
Intuitively, stream (2) generates coarse instance-level masks that are refined by pixel-accurate background masks from stream (1). Specifically, pixels where rigid background and instance predictions disagree are not used for rigid body fitting below, and marked as wrong prediction in evaluation.

\noindent{\bf Rigid body scene flow:}\quad
Given rigid motion segmentations, scene flow of the background and N rigid instances are parameterized by their rotations, translations $\{({\bf R}_i,{\bf T}_i),i=0,\cdots,N\}$ and the first frame 3D scene points ${\bf P_0}$.  To find the best fit of rotations and up-to-scale translations, we estimate and decompose essential matrices over flow correspondences with a least median of squares estimator~\cite{hartley2003multiple}. To obain a more reliable 3D reconstruction than back-projected monocular depth, we triangulate flow using rigid motion estimations for each rigid body, and determine their scales by aligning each triangulated depth map to monocular depth inputs with robust least squares~\cite{sun2010secrets}. With the above parameterization, the second frame coordinates are computed as,
\begin{equation}\footnotesize
\label{eq:rbsf}
{\bf P_1} = \textstyle \sum_{i=0}^N{\bf S}_i ({\bf R_i}{\bf P_0} +{\bf T_i}),
\end{equation}
where ${\bf S}_i$ is a one-hot rigid motion segmentation vector.

\begin{figure*}
\centering
\includegraphics[width=0.9\linewidth]{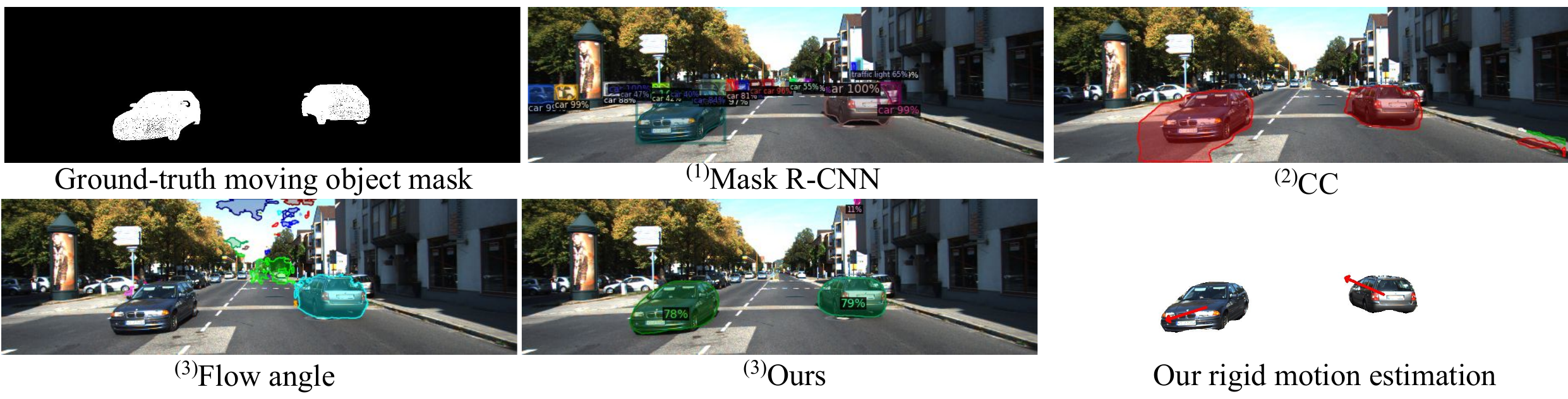}
\vspace{-5pt}
\caption{Comparison on KITTI-SF, image 137. The prefix of each method indicates the test-time inputs: $^{(1)}$Single frame. $^{(2)}$Multi-frame with appearance features. $^{(3)}$Multi-frame without appearance. Our best appearance-based segmentation baseline, $^{(1)}$Mask R-CNN~\cite{he2017mask} detects all the moving vehicles, but also reports parked cars in the background. $^{(2)}$CC~\cite{ranjan2019competitive} correctly detects moving cars but also reports the edge of the road as moving objects. $^{(3)}$Geometric segmentation algorithm ~\cite{bideau2016s,Wulff:CVPR:2017} fails on the approaching car due to the colinear motion ambiguity, and reports false positives at the background due to the noisy flow estimation. In contrast, $^{(3)}$our method correctly segments both the moving vehicles and the rigid background. Rigid motions are estimated within each mask and applied to depth and scene flow estimation.}
\label{fig:moseg-kitti}
\vspace{-10pt}
\end{figure*}

\section{Experiments}
Our method is quantitatively compared with state-of-the-art rigidity estimation algorithms on KITTI and Sintel in Sec.~\ref{sec:exp1}, and then applied to the depth and scene flow estimation tasks in Sec.~\ref{sec:exp2}. In Sec.~\ref{sec:exp3} we conclude with an ablation study.
 
 \begin{table}[!t]
    \caption{
    Rigidity estimation on KITTI (K) and Sintel (S) without fine-tuning. $^{(1)}$Single frame. $^{(2)}$Multi-frame with appearance features. $^{(3)}$Multi-frame without appearance. The best result under each metric (IoU in \%) is bolded. $^*$:For methods only estimating background masks, we use connected components to obtain object masks. $^\ddagger$:Methods trained on target dataset. MR-Flow-S (K) is trained on KITTI, and MR-Flow-S (S) is trained on Sintel.}
    \vspace{-10pt}
    \label{tab:moseg-test}
    \small
    \centering
    \begin{tabular}{lrlllllll}
	\toprule
  & Method & K: obj $\uparrow$ & K: bg $\uparrow$& S: bg $\uparrow$\\
\midrule
\multirow{4}{*}{\shortstack{(1)}}
& Mask R-CNN~\cite{wu2019detectron2}	         & 88.20  & 96.42    & 81.98\\
& U$^2$ (Saliency)~\cite{Qin_2020_PR}			 & 64.80$^{*}$ & 93.34	 & 82.01\\
& MR-Flow-S (K) ~\cite{Wulff:CVPR:2017}          & 75.59$^{*}$ & 94.70$^{\ddagger}$        & 76.11\\
& MR-Flow-S (S) ~\cite{Wulff:CVPR:2017}          & 11.11$^{*}$  & 84.72    & 92.64$^{\ddagger}$\\
\midrule
\multirow{5}{*}{\shortstack{(2)}}
& FSEG~\cite{jain2017fusionseg}	         & 85.08$^{*}$       & 96.27     & 80.22\\
& MAT-Net~\cite{zhou2020motion}	                 & 68.40$^{*}$       & 93.08     & 77.95\\
& COSNet~\cite{lu2019see}				         & 66.67$^{*}$       & 93.03     & 80.86\\
& CC~\cite{ranjan2019competitive}               & 50.87$^{*}$        & 85.50   &  \xmark\\
& RTN~\cite{Lv18eccv}					        & 34.29$^{*}$        & 84.44   & 64.86\\
\midrule
\multirow{4}{*}{\shortstack{(3)}}
& FSEG-Motion~\cite{jain2017fusionseg}          & 61.29 & 89.41&78.25\\
& CC-Motion~\cite{ranjan2019competitive}&42.99&74.06&  \xmark\\
& Flow angle~\cite{bideau2016s,Wulff:CVPR:2017} & 25.83 &85.52&74.23\\
&Ours		                                    & {\bf 90.71}    & {\bf 97.05}       & {\bf 86.72}\\
\bottomrule
\end{tabular}
\vspace{-10pt}
\end{table}
 
\noindent{\bf Dataset:}\quad 
We use KITTI-SF (sceneflow) and Sintel for quantitative analysis. KITTI-SF~\cite{Geiger2012CVPR,Menze2015CVPR} features an urban driving scene with multiple rigidly moving vehicles. Sintel~\cite{butler2012naturalistic} is a synthetic movie dataset that features a highly dynamic environment. It contains viewpoints and objects (such as dragons) that are rare in existing datasets. KITTI provides ground-truth segmentation masks for the rigid background and moving car instances, where the remaining dynamic objects (such as pedestrians) are manually removed. For Sintel, computing rigid instances masks is an ill-posed problem since most objects are nonrigid. Instead, we obtain ground-truth rigid background segmentation from MR-Flow~\cite{Wulff:CVPR:2017}. Both datasets also provide ground-truth depth and scene flow as well as camera intrinsics. 

\noindent{\bf Setup {I} (monocular):}\quad 
In the monocular setup, depth and scene flow is predicted from two consecutive frames. We use MiDaS~\cite{Ranftl2019}, a state-of-the-art monocular depth estimator to acquire imprecise, up-to-scale depth of the first frame as inputs. The remaining networks are trained without target domain data: optical flow and optical expansion networks are trained using FlyingChairs, SceneFlow, VIPER, and HD1K~\cite{DFIB15,kondermann2016hci,MIFDB16,richter2017playing}; the rigid motion segmentation network is trained on SceneFlow~\cite{MIFDB16} with other networks being fixed. Finally, the performance of rigid motion segmentation and scene flow is evaluated on KITTI-SF and Sintel.

\noindent{\bf Setup {II} (stereo):}\quad 
Our method is able to take advantage of reliable depth sensors, such as stereo cameras, to produce better segmentation and scene flow estimation. Different than the monocular setup, two consecutive stereo pairs are used as inputs. Specifically, we use GA-Net~\cite{zhang2019ga} to acquire metric depth information from the first stereo pair as inputs to the segmentation network. To take advantage of accurate depth estimation in the first frame, we refine the estimated rigid transformations by solving a Perspective-n-Point problem given first frame depth and flow using non-linear optimization. To estimate optical flow and optical expansion, we use an off-the-shelf network pre-trained on KITTI-SF~\cite{yang2020upgrading}. To train the rigid motion segmentation network, we mix KITTI-SF and SceneFlow. The performance of stereo scene flow is reported on the KITTI benchmark.

\subsection{Two-frame Rigid Motion Segmentation}
\label{sec:exp1}

\noindent{\bf Metrics:}\quad Following prior works, we compute background IoU~\cite{Lv18eccv, ranjan2019competitive} for rigid background segmentation and object F-measure~\cite{dave2019towards} for rigid instance segmentation. Only the rigid background segmentation metric is reported on Sintel due to the lack of rigid bodies ground-truth rigid motion segmentation masks.

\noindent{\bf Baselines:}\quad We group baselines according to test inputs. 

\noindent$^{(1)}${\bf Single frame methods}. Mask R-CNN with ResNeXt-101+FPN backone is the most accurate model on MSCOCO provided by Detectron2~\cite{he2017mask,lin2014microsoft,wu2019detectron2,xie2017aggregated}; U$^2$Net~\cite{Qin_2020_PR} is a state-of-the-art salient object detector; and MR-Flow-S~\cite{Wulff:CVPR:2017} is a semantic rigidity estimation network fine-tuned separately on KITTI and Sintel. 

\noindent$^{(2)}${\bf  Multi-frame with appearance features}. FusionSeg~\cite{jain2017fusionseg} is a two-stream architecture that fuses the appearance and optical flow features, and we provide SOTA optical flow on KITTI and Sintel as motion input.  COSNet~\cite{lu2019see} and MATNet~\cite{zhou2020motion} are SOTA video objection segmentation methods on DAVIS~\cite{Perazzi2016}. CC~\cite{ranjan2019competitive} combines ``flow-egomotion consensus score'' (similar to our epipolar costs) with a foreground probability regressed from five consecutive frames, which is then thresholded to obtain the background mask. RTN~\cite{Lv18eccv} uses a CNN to predict rigid background masks given two RGBD images. For Sintel, we use the ground-truth depth as input; for KITTI, since the ground-truth depth is sparse, we use MonoDepth2~\cite{monodepth2} instead.

\noindent$^{(3)}${\bf Two-frame without appearance}.
We separately evaluate the motion stream of FSEG and the flow-egomotion consensus results of CC. Following prior work~\cite{bideau2016s,Wulff:CVPR:2017}, we implement a classic motion segmentation pipeline that combines the motion angle and motion residual criteria. 

Besides CC, RTN, and the classic pipeline, all baselines are trained or pre-trained on large-scale manually annotated segmentation datasets that contain common objects and scenes, while ours is not.

\noindent{\bf Performance analysis:}\quad
We show qualitative comparison in Fig.~\ref{fig:moseg-kitti} and report results in Tab.~\ref{tab:moseg-test}. On KITTI, our method outperforms the most accurate baseline, Mask R-CNN, in terms of both rigid instance segmentation and background segmentation. Although Mask R-CNN is trained on common scenes (including driving), it cannot tell whether an object is moving or static, similar to other single-frame methods. Therefore, our method compares favorably to Mask R-CNN on rigid motion segmentation task. On Sintel, our method outperforms all the baselines except MR-Flow-S (S), which uses the first half of all Sintel sequences for training. If we compare to MR-Flow-S (K), which is not fine-tuned on Sintel, our method is better. Finally, among the motion-based segmentation methods, our method is the best on both datasets, because of our robustness to degenerate motion configurations as well as noisy flow inputs.

\begin{table}
   \caption{Monocular depth and scene flow results on KITTI (K) and Sintel (S). D1: first frame disparity (inverse depth) error. SF: scene flow error (\%). The best result is underlined, and further bolded if not trained on the target domain data. On Sintel, we evaluate on 330 frame pairs with average flow magnitude greater than 5 pixel.}
      \small
    \centering
   \begin{tabular}{rcccccccc}
	\toprule
    Method & K: D1 $\downarrow$ & K:SF $\downarrow$ &S: D1 $\downarrow$& S:SF $\downarrow$\\
\midrule
 CC~\cite{ranjan2019competitive}                            &36.20  &51.80  &\xmark&\xmark\\
 SSM~\cite{Hur:2020:SSM}                                                &31.25  &47.05   &\xmark&\xmark\\ 
 Mono-SF~\cite{brickwedde2019mono}                                                    &\underline{16.72}  & \underline{21.60}&\xmark&\xmark\\
 \midrule

 MiDaS+OE~\cite{yang2020upgrading}			     &37.27	&44.87 &49.89 & 55.43\\
 MiDaS+Mask                                      & 17.33  &22.47 &39.60 &47.40 \\
 MiDaS+Ours                                      & {\bf 16.98} &{\bf 22.19} &\underline{\bf 38.29} & \underline{\bf 46.05}\\

\bottomrule
\label{tab:depth}
\end{tabular}    
\vspace{-20pt}
\end{table}

\begin{table}
  \caption{Stereo scene flow results on KITTI benchmark. D1 and D2: first and second frame disparity error. Fl: optical flow error. SF: scene flow error. Metrics are errors in percentage and top results are bolded. $^*$First frame disparity is not refined by our method.}
  \small
    \centering
    \begin{tabular}{rcccccccc}
	\toprule
    Method & D1$^{*}$ $\downarrow$&D2 $\downarrow$&Fl $\downarrow$&SF $\downarrow$\\
\midrule
PRSM~\cite{Vogel2015IJCV}			& 4.27	&6.79 	& 6.68	& 8.97\\
OpticalExp~\cite{yang2020upgrading}	& 1.81	&4.25 	&6.30	& 8.12\\
DRISF~\cite{Ma2019CVPR}				& 2.55	&4.04 	&4.73	& 6.31\\
Ours Mask R-CNN                     & 1.89  & 3.42      &4.26    & 5.61\\
Ours Rigid Mask				        & 1.89	& {\bf 3.23}&{\bf 3.50}&{\bf 4.89}\\
\bottomrule
\label{tab:sf-test}
\end{tabular}
\vspace{-20pt}
\end{table}

\subsection{Rigid Depth and Scene Flow Estimation}
\label{sec:exp2}
The rigid motion masks are then applied to the two-frame depth and scene flow estimation task under both monocular and stereo setups. In both cases, we fit a 3D rigid transformation for pixels within each rigid body mask, which is used to update the depth and scene flow estimation.

\noindent{\bf Metrics:} To evaluate depth and scene flow estimation performance on KITTI and Sintel, we report disparity error on both frames (D1, D2), optical flow error (Fl) and scene flow error (SF) following KITTI~\cite{Menze2015CVPR}. In the monocular setup, to remove the overall scale ambiguity, we take an extra step to align the overall scale of the predictions to the ground-truth with their medians~\cite{Ranftl2019, wang2018learning}.

\begin{figure*}
\centering
\includegraphics[width=0.9\linewidth]{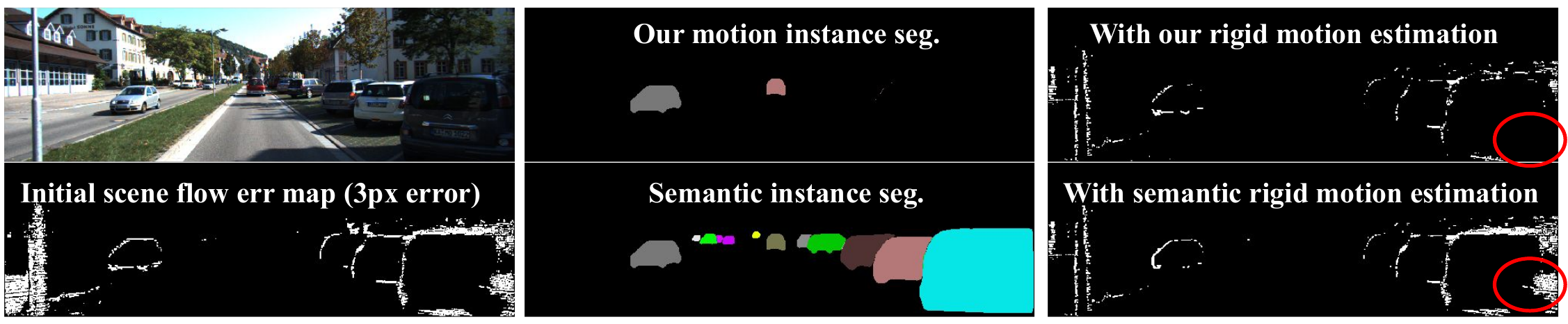}
\setlength{\belowcaptionskip}{-10pt}
\caption{Rigidity vs semantic-based segmentation for instance scene flow. Given instance segmentation masks, scene flow can be optimized by fitting rigid body transforms within each mask. While semantic segmentation fails to improve scene flow estimation on the parked cars (in red circle), our rigid motion mask groups the parked car together with the rigid background and successfully reduces the scene flow error.}
\label{fig:sf}
\end{figure*}

\noindent{\bf Setup I (monocular):} We compare against state-of-the-art monocular scene flow baselines. {\bf CC}~\cite{ranjan2019competitive} and {\bf SSM}~\cite{Hur:2020:SSM} are representative methods for self-supervised monocular depth and scene flow estimation that does not make use of segmentation priors at inference time. {\bf Mono-SF}~\cite{brickwedde2019mono} trains a monocular depth network with KITTI ground-truth, and solve an optimization problem given semantic instance segmentation provided by Mask R-CNN. The above three methods are trained on KITTI and the results are taken from their papers. {\bf OE} (optical expansion)~\cite{yang2020upgrading} learns to predict relative depth from dense optical expansion, which together with optical flow, directly yields 3D motion. It is trained on the synthetic SceneFlow dataset, and we use MiDaS to provide the scale. We also implement a baseline ({\bf MiDaS+Mask}) that predicts instance segmentation masks by Mask R-CNN, and follows the same rigid body fitting procedure as ours.

\noindent{\bf Performance analysis:}\quad 
We report results on KITTI-SF and Sintel in Tab.~\ref{tab:depth}. First, it is noted our method reduces the disparity error of MiDaS by more than 50\% on KITTI, and 20\% on Sintel. Compared to OE, which uses the same monocular depth input as ours, we are better in all metrics.  (SF: 22.19\% vs 44.87\%), which demonstrates the effectiveness of our rigid motion mask.  Our method also outperforms the other methods that do not use segmentation priors (CC and SSM). Compared to Mono-SF, which is trained with ground-truth KITTI depth maps, and uses a semantic segmentation prior, our method is slightly worse on KITTI. Compared to Midas-Mask, our method is strictly better on both KITTI and Sintel, indicating the benefit of using our rigid motion masks versus appearance-based masks.

\noindent{\bf Setup II (stereo):}\quad 
Our method segments rigid motions based on two-frame rigidity and fits rigid body transformations over depth-flow correspondences, which is used to update the second frame depth and flow estimations. Among the baselines, {\bf OE}~\cite{yang2020upgrading} uses the same architecture (as in the monocular setup) fine-tuned on KITTI to upgrade optical flow to 3D scene flow. Same as ours, GA-Net stereo and VCN optical flow are used as inputs. {\bf PRSM}~\cite{Vogel2015IJCV} segments an image into superpixels, and fits rigid motions to estimate piece-wise rigid scene flow. Given semantic instance segmentation~\cite{he2017mask}, depth, and optical flow, {\bf DRISF}~\cite{Ma2019CVPR} casts scene flow estimation as an energy minimization problem and finds the best rigid transformation for each \textit{semantic} instance. The key difference between our method and DRISF is that we use rigid motion segmentation masks.

\noindent{\bf Performance analysis:}\quad 
As reported in Tab.~\ref{tab:sf-test}, our method demonstrates state-of-the-art performance on KITTI scene flow benchmark (SF: 4.89 vs 6.31). If we replace the segmentation masks with semantic instance segmentation, i.e., Mask R-CNN, the performance drops noticeably (SF: 4.89\% to 5.61\%). As illustrated in Fig.~\ref{fig:sf}, our method successfully groups the static objects (e.g. parked cars) with the rigid background, which effectively improves scene flow accuracy by optimizing the whole background as one rigid body, while semantic instance segmentation methods fail to do so.

\begin{table} 
    \caption{Diagnostics of rigid body motion segmentation on KITTI-SF. Dignostics in the second group are sequential.}
    \small
    \centering
    \begin{tabular}{lcccccccc}
	\toprule
Method & K: obj $\uparrow$ & K: bg $\uparrow$& S: bg $\uparrow$\\
\midrule
Reference			    	& {\bf 89.53}	& {\bf 97.22} & {\bf 84.63}\\
$^{(1)}$w/o cost maps		        & 88.66 & 96.59     & 76.81\\
$^{(2)}$w/o uncertainty		        & 85.09 & 95.72     & 77.25\\
$^{(3)}$w/o monocular depth	        & 84.46 & 94.84     & 76.14\\
$^{(4)}$w/o expansion (MoA~\cite{bideau2018moa}) 		&81.28& 95.50& 77.00\\
$^{(5)}$w/o learning~\cite{bideau2016s,Wulff:CVPR:2017} & 25.83 &85.52&74.23\\
\bottomrule
\label{tab:aba}
\end{tabular}
\vspace{-30pt}
\end{table}

\subsection{Diagnostics}
\label{sec:exp3}
We ablate critical components of our approach and retrain networks. Results are shown in Tab.~\ref{tab:aba}. We validate the design choices of using $^{(1)}$explicitly computed rigidity cost-maps inputs, $^{(2)}$uncertainty estimation inputs, $^{(3)}$monocular depth inputs, $^{(4)}$optical expansion that upgrades 2D optical flow to 3D, and $^{(5)}$ our rigid motion segmentation network. $^{(1)}$Removing rigidity cost-maps leads to a slight drop of accuracy on KITTI, and a significant drop on Sintel (84.63\% to 76.81\%). This indicates the cost map features are crucial for Sintel, possibly due to complex camera and object motion configurations, in which cases explicit geometric priors are helpful. $^{(2)}$Removing uncertainty inputs leads to a noticible drop of performance on KITTI (88.66\% to 85.09\%). We posit uncertainty estimation contains rich information about motion distribution, and is therefore useful for segmentation. $^{(3)}$Further removing monocular depth inputs leads to an accuracy drop on all metrics, especially on KITTI, which shows the importance of using depth cues to deal with collinear motions in autonomous driving scenes. $^{(4)}$After further removing optical expansion, our method degrades to MoA-Net~\cite{bideau2018moa}. The performance drops noticeably on KITTI rigid instance segmentation metric (84.46\% to 81.28\%), which indicates optical expansion is useful for segmenting foreground objects. $^{(5)}$Lastly, if we directly apply the rigidity cost maps with manually-tuned thresholds to decide the background region without the neural architecture and learning, the method becomes worse in all metrics due to the lost of robustness to noisy inputs and degenerate motion.

\section{Conclusion}
We investigate the problem of two-frame rigid body motion segmentation in an open environment. We analyze the degenerate cases in geometric motion segmentation and introduce novel criteria and inputs to resolve such ambiguities. We further propose a modular neural architecture that is robust to noisy observations as well as different motion types, which demonstrates state-of-the-art performance on rigid motion segmentation, depth and scene flow estimation tasks. 

{\bf Acknowledgements:} This work was supported by the CMU Argo AI Center for Autonomous Vehicle Research.

{\small
\bibliographystyle{ieee_fullname}
\bibliography{egbib}
}

\clearpage
\newpage

\section{Appendix}
\subsection{Rigidity cost maps}
In Sec.~3.2, we briefly motivated the design choice of using rigidity cost-maps inputs, and we expand the particular cost functions here. Given motion correspondences $({\bf p_0},{\bf p_1}) \in R^2$, camera intrinsics $({\bf K_0}, {\bf K_1})$, and camera motion ${\bf R_c} \in SO(3)$, ${\bf T_c} \in R^3$, we construct four geometric motion cost maps that are tailored to particular motion configurations, including 1) an epipolar cost, 2) a homography cost, 3) a 3D P+P cost, and 4) a depth contrast cost. 

\noindent\textbf{1) Epipolar costs} are applied to detect general moving objects, computed as the classic Sampson error~\cite{hartley2003multiple} per-pixel. We include it here for completeness:
\begin{equation}
c_{\text {epi}} = \frac{({\bf \tilde{p}_1}^T{\bf F}{\bf \tilde{p}_0})^2}{({\bf F}{\bf \tilde p_0})_1^2+({\bf F}{\bf \tilde p_0})_2^2+({\bf F}^T{\bf \tilde p_1})_1^2+({\bf F}^T{\bf \tilde p_1})_2^2 + \epsilon},
\end{equation}
where ${\bf F} = {\bf K_1}^{-T}{\bf R}[{\bf t}]_{\times}{\bf K_0}^{-1}$ is the fundamental matrix and $({\bf \tilde{p}_0},{\bf \tilde{p}_1})$ are motion correspondences in the homogeneous coordinates. $\epsilon=10^{-9}$ is a constant value added for numerical stability.

\begin{figure*}
    \centering
    \includegraphics[width=0.8\linewidth]{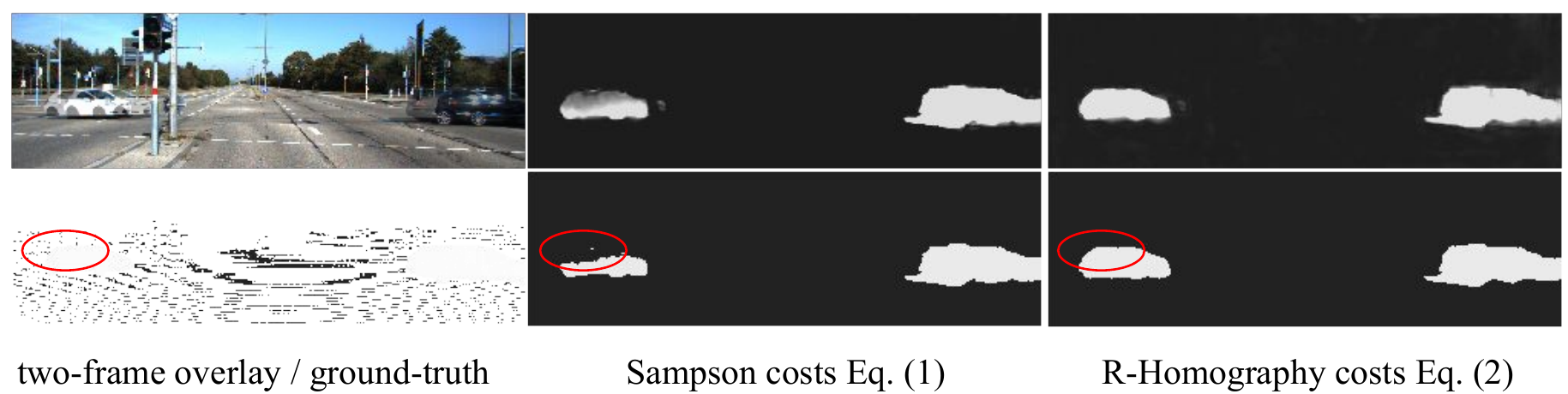}
    \caption{Epipolar costs vs homography costs. {\bf Top}: Gray-scale costs values. {\bf Bottom}: Binary segmentations after thresholding the costs. This scene features two moving foreground cars and  a static camera that causes motion degeneracy in epipolar geometry (e.g., the low-cost but moving region in the Sampson cost map, marked by the red circle). In such cases, epipolar line is not well-defined, and the homography model is more suitable for motion segmentation.}
    \label{fig:hom-costs}
    \end{figure*}

\noindent\textbf{2) Homography costs} are applied to deal with motion degeneracies in epipolar geometry~\cite{torr1999problem}, when it becomes difficult to estimate camera translation, but not rotation~\cite{cai2019equivalent}. A visual comparison between epipolar costs and homography costs can be found in~Fig.~\ref{fig:hom-costs}. The homography cost is implemented as per-pixel symmetric transfer error~\cite{dubrofsky2009homography} with regard to the rotational homography, ${\bf H_R}={\bf K_0}{\bf R_c}{\bf K_1}^{-1}$,
\begin{equation}
c_{\text {hom}} = d({\bf \tilde p_0}, {\bf H_R}{\bf \tilde p_1})^2+d({\bf \tilde p_1}, {\bf H_R}^{-1}{\bf \tilde p_0})^2,
\end{equation}
where $d(\cdot,\cdot)$ is the Euclidian image distance between two points. 

\begin{figure*}
    \centering
    \includegraphics[width=\linewidth]{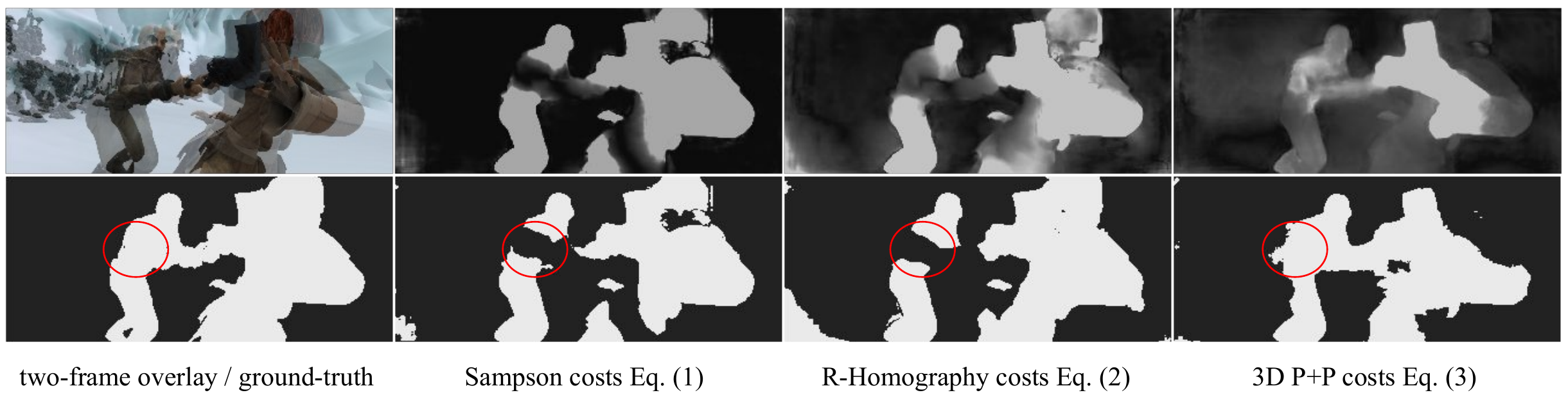}
    \caption{3D P+P costs. {\bf Top}: Gray-scale costs. {\bf Bottom}: Binary segmentations after thresholding the costs. This scene contains a moving camera and nonrigid dynamic objects, where pixels that move along the epipolar line are not recoverable under classic motion segmentation criteria (e.g., the low-cost but moving region in the Sampson cost map, marked by the red circle). In such cases, our 3D P+P cost is more suitable.}
    \label{fig:3d-costs}
    \end{figure*}

\noindent\textbf{3) 3D P+P costs} are applied to detect the coplanar motion, where points are moving along the epipolar line (not detectable by the epipolar costs, as analyzed in Sec. 3.1. Our 3D P+P cost is extended from the 2D residual error of~\cite{bideau2016s},
\begin{equation}
c_{\text {3D}} = ||{\bf \tilde{T}_{sf}}||\cdot|\sin\beta|,
\end{equation}
where $\beta = |\angle({\bf \tilde{T}_{sf}},-{\bf T_c})|$ is the measured angle between the normalized scene flow ${\bf \tilde{T}_{sf}}$ (as computed through optical expansion using the method of~\cite{yang2020upgrading}) and negative camera translation ${\bf -T_c}$, capped to $\frac{\pi}{2}$. A visual comparison is shown in Fig.~\ref{fig:3d-costs}.

\begin{figure*}
    \centering
    \includegraphics[width=\linewidth]{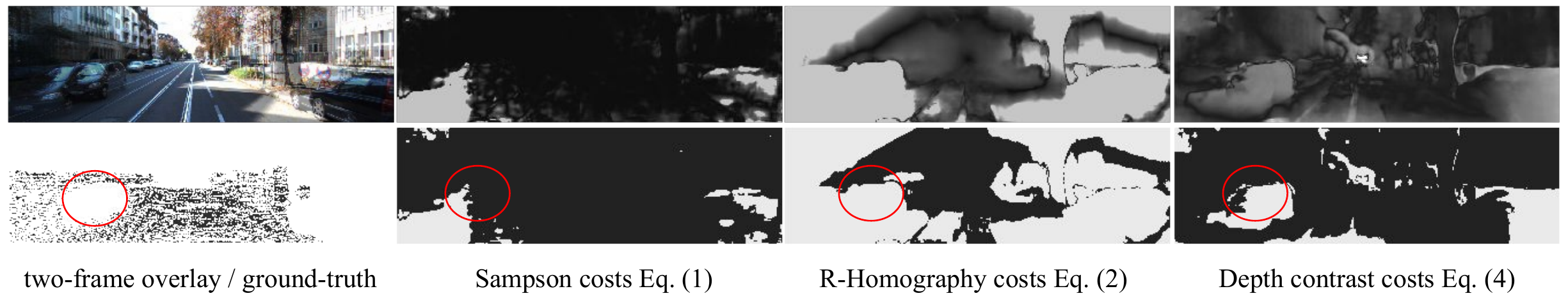}
    \caption{Depth contrast costs. {\bf Top}: Gray-scale costs. {\bf Bottom}: Binary segmentations after thresholding the costs. This scene contains a moving camera and a rigid body (car) moving along the negative direction of camera translation, which is not recoverable under classic motion segmentation criteria (e.g., the low-cost but moving region in the Sampson cost map, marked by the red circle) as well as the 3D P+P cost. In such cases, our depth contrast cost is more suitable.}
    \label{fig:depth-costs}
    \end{figure*}

\begin{figure*}
    \centering
    \includegraphics[width=\linewidth]{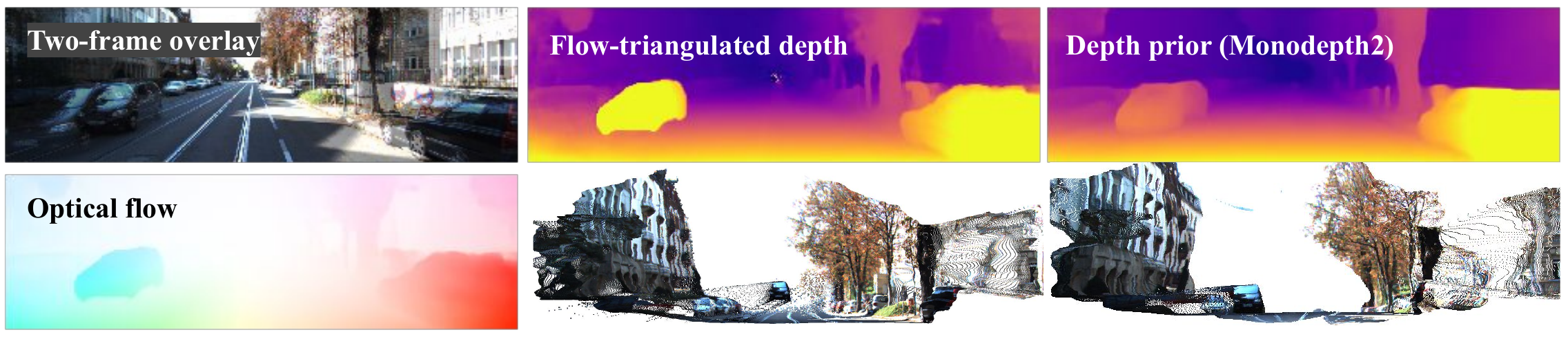}
    \caption{Flow-triangulated depth vs monocular depth prior. The flow-triangulated depth $Z^{flow}_{0}$ (middle) is the triangulation of motion correspondences assuming overall rigidity. The monocular depth prior $Z^{prior}_{0}$ (right) can be represented by a data-driven monocular depth network~\cite{Ranftl2019}. In this example, the left vehicle is moving opposite to the camera translation direction, and cannot be detected by epipolar constraints, as shown in Fig.~2 of the main text. However, it appears abnormal (floating above the ground) in the flow-triangulated reconstruction. To detect such collinearly moving objects, we globally align $Z^{prior}_{0}$ to $Z^{flow}_{0}$ by a scale factor $\gamma$ with robust least squares~\cite{sun2010secrets}, which reveals the floating (moving) car that is inconsistent with the monocular depth prior.}
    \label{fig:depth-contrast}
    \end{figure*}

\noindent\textbf{4) Depth contrast costs} are applied to address the colinear motion ambiguity, where points are moving opposite to the camera translation direction in 3D, and therefore not detectable by the above costs, as shown in Fig.~\ref{fig:depth-costs}. The depth contrast cost is implemented as:
\begin{equation}
c_{\text {depth}} = |\log(\frac{Z^{\text {flow}}}{\gamma Z^{\text {prior}}})|,
\end{equation}
where the flow-triangulated depth $Z^{flow}_{0}$ can be computed efficiently using midpoint or DLT triangulation algorithm~\cite{hartley2003multiple}, the monocular depth prior $Z^{prior}_{0}$ can be represented by a data-driven monocular depth network~\cite{monodepth2}, and the scale factor $\gamma$ that globally aligns $Z^{prior}_{0}$ to $Z^{flow}_{0}$ can be determined by robust least squares~\cite{sun2010secrets}. A visual comparison between the flow-triangulated depth and monocular depth prior is shown in Fig.~\ref{fig:depth-contrast}.

\subsection{Training details}
The details for training optical flow, optical expansion and rigid motion segmentaion networks are shown in Tab.~\ref{tab:hp}. 

\begin{table}
    \caption{Details for network training. C: FlythingChairs~\cite{DFIB15}. T: FlythingThings~\cite{MIFDB16}. SF: SceneFlow~\cite{MIFDB16}. V: VIPER~\cite{richter2017playing}. The optical flow network is trained sequentially on C, T, and C+SF+V.}
    \footnotesize
    \centering
    \begin{tabular}{lr}
	\toprule
 Parameter  & Value \\
 \midrule
  \multicolumn{2}{c}{\bf Optical flow}\\
 Network architecture	& VCN~\cite{yang2019volumetric}\\
 Optimizer          	&   Adam~\cite{kingma2014adam}\\
 Learning rate             &   $1\times 10^{-3}$\\
 Batch size / iterations on C	& 16 image pairs / 70k\\
 Batch size / iterations on T	& 16 image pairs / 70k\\
 Batch size / iterations on C+SF+V	& 12 image pairs / 70k\\
 \midrule
 \multicolumn{2}{c}{\bf Optical expansion}\\
 Network backbone	& U-Net~\cite{ronneberger2015u,yang2020upgrading}\\
 Optimizer          	&   Adam~\cite{kingma2014adam}\\
 Learning rate             &   $1\times 10^{-3}$\\
 Batch size / iterations on SF	& 12 image pairs / 70k\\
 \midrule
 \multicolumn{2}{c}{\bf Rigid motion segmentation}\\
 Network backbone	& U-Net+DLA-34~\cite{ronneberger2015u,yang2020upgrading,yu2018deep,zhou2019objects}\\
 Optimizer          	&   Adam~\cite{kingma2014adam}\\
 Learning rate             &   $5\times 10^{-4}$\\
 Batch size / iterations on SF	& 12 image pairs / 70k\\
 \bottomrule
\label{tab:hp}
\end{tabular}
\vspace{-10pt}
\end{table}

\subsection{Details of rigid body scene flow}
In Sec.~3.2, we describe rigid body scene flow that (1) fits 3D rigid motions per rigid body, and (2) updates depth as well as flow measurements. More details are provided here. 

Overalll, our goal is to select \emph{high-quality flow correspondences} for model fitting, and update the rigid bodies with \emph{large enough motion}. To do so, we first define ``valid pixels'' as pixels with flow confidence (in range 0-1, estimated by VCN~\cite{yang2019volumetric}) greater than 0.5. During fitting, we use flow correspondences of valid pixels from each rigid motion mask to fit a essential matrix through a least median of squares estimator~\cite{rousseeuw1984least}. Then, each essential matrix is decomposed to four rotations and up-to-scale translations, where only one is feasible through cheirality check~\cite{hartley2003multiple}. To determine the scale of translation, we triangulate flow correspondences at vailid pixels and align it with the initial depth input by a scale factor through RANSAC~\cite{fischler1981random}. To take advantage of accurate depth estimation in the stereo case, we refine the estimated rigid transformations by solving a Perspective-n-Point problem given first frame depth and flow that minimizes re-projection errors with Levenberg–Marquardt algorithm~\cite{hartley2003multiple}.

Finally, we update depth and flow estimations according to the estimated 3D rigid motions. Rigid bodies whose average parallax flow magnitude (defined as ``rectified'' optical flow after rotation removal in Sec.3.1) is lower than 4px, or has fewer than 30\% valid pixels are not updated.

\begin{table*}
  \caption{Ablation study of stereo scene flow on KITTI-SF images. D1 and D2: first and second frame disparity error. Fl: optical flow error. all: evaluated on all pixels. fg: evaluated on foreground pixels only. SF: scene flow error. $\Delta$: percentage of error reduction after refinement. $^*$First frame disparity does not change during refinement.}
  \small
    \centering
    \begin{tabular}{rcccccccccccc}
	\toprule
	&\multicolumn{2}{c}{$^*$D1 (\%)} & \multicolumn{2}{c}{D2 (\%)} & \multicolumn{4}{c}{Fl (\%)} & \multicolumn{4}{c}{SF (\%)} \\
	\cmidrule(lr){2-3} \cmidrule(lr){4-5} \cmidrule(lr){6-9} \cmidrule(lr){10-13}
    Method & \emph{all} $\downarrow$&\emph{fg} $\downarrow$ &\emph{all} $\downarrow$&\emph{fg} $\downarrow$ &\emph{all} $\downarrow$ & $\Delta$-all $\uparrow$&\emph{fg} $\downarrow$& $\Delta$-fg $\uparrow$ &\emph{all} $\downarrow$&$\Delta$-all $\uparrow$&\emph{fg} $\downarrow$&$\Delta$-fg $\uparrow$\\
\midrule
Baseline OE~\cite{yang2020upgrading} 	& 1.41&0.76 & 2.45&{\bf 0.91} & 4.02&0 &2.50&	0 & 5.12&0&3.07&0 \\
Ours Mask R-CNN                 				& 1.41&0.76 & 2.11&1.99 & 3.53& 12.1 &4.34 & -73.6 &4.02 & 21.5 &4.86 & -36.8\\
Ours Rigid Mask							& 1.41&0.76	& {\bf 2.04}&1.05 & {\bf 3.32}& {\bf 17.4} & {\bf 2.16}	&{\bf 15.7}  &{\bf 3.86}&{\bf 24.6}&{\bf 2.78} &{\bf 10.4}\\
\bottomrule
\label{tab:sf-aba}
\end{tabular}
\vspace{-20pt}
\end{table*}

\subsection{Ablation study of rigid body scene flow}
We study the effect of rigid motion parameterization for scene flow estimation and report results on 200 images of KITTI-SF as shown in Tab.~\ref{tab:sf-aba}. Without rigid motion parameterization, our method is equivalent to optical expansion~\cite{yang2020upgrading}, which upgrades 2D flow fields to 3D, but does not refine the first frame disparity as well as optical flow. In contrast, the proposed method reduces the overall scene flow error by 24.6\% through rigid body refinement. Replacing the proposed rigid motion masks with appearance-based masks produced by Mask R-CNN leads to a noticible accuracy drop. Our rigid body parameterization also lead to constant improvement of scene flow accuracy for both foreground and background regions.

\subsection{Qualitative comparison}

\begin{figure*}
    \centering
    \includegraphics[width=\linewidth]{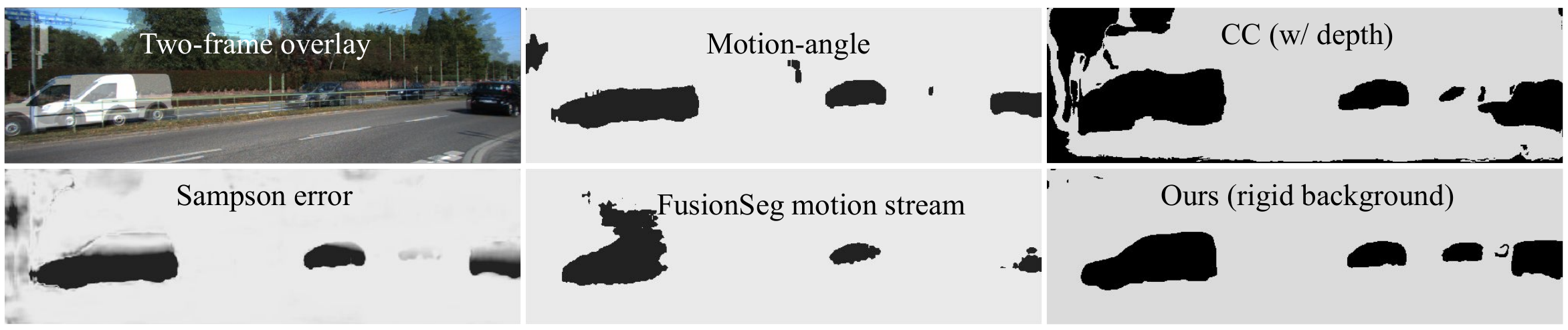}
    \caption{Illustration of coplanar motion ambiguity on KITTI. Rigid background are indicated by the white color. Points moving along the epipolar line, for example the roof of the cars, yields small Sampson error, and therefore are estimated as background in classic geometric pipelines. We make use of optical expansion, which reveals the relative depth change, to resolve such ambiguity. Compared to prior motion-based segmentation method, ours is more robust to noise.}
    \label{fig:kitti}
    \end{figure*}

\begin{figure*}
    \centering
    \includegraphics[width=0.85\linewidth]{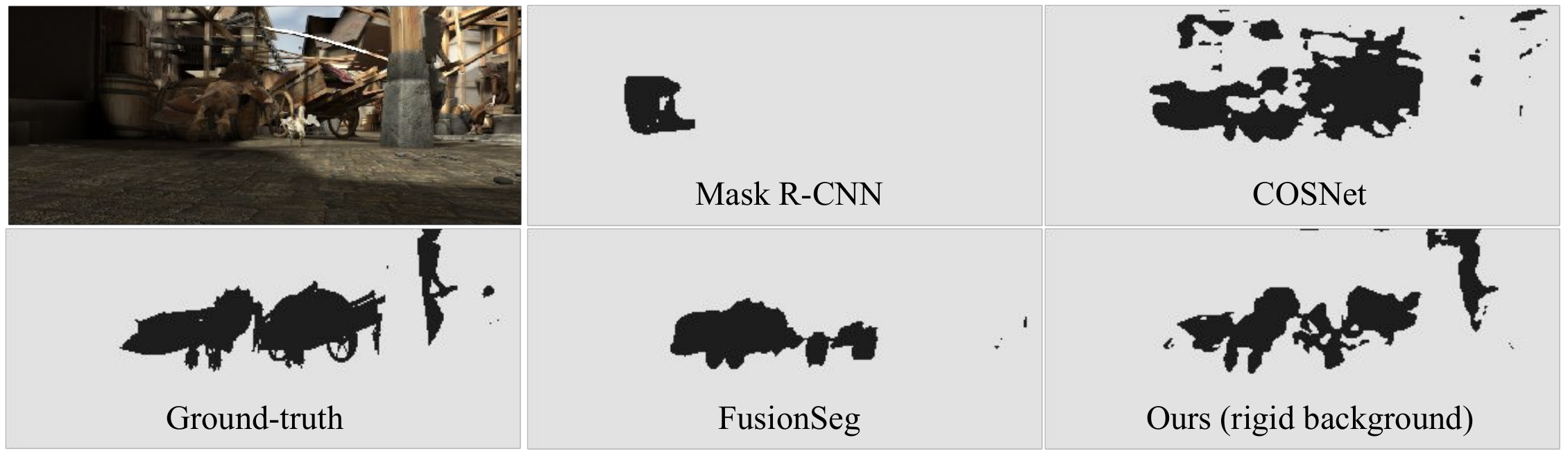}
    \caption{Results on Sintel market sequence. Prior single frame or video motion segmentation methods fail due to unusual view-point (viewing from the ground) and never-before-seen objects (dragon, wood carts). Our method accurately segments novel moving objects.}
    \label{fig:sintel}
    \end{figure*}

We provide additional visual comparison with prior approaches on KITTI and Sintel in Fig.~\ref{fig:kitti} and Fig.~\ref{fig:sintel}. Compared to appearance-based methods for segmenting rigid motions, our method is able to correctly segment the static objects as part of the rigid background, and generalizes to novel appearance. Compared to geometric motion segmentation methods, our method is more robust to degenerate motion configurations and noisy flow as well as camera inputs.

\end{document}